# THERMAL HAND IMAGE SEGMENTATION FOR BIOMETRIC RECOGNITION

Xavier Font-Aragones (1), Marcos Faundez-Zanuy (1), Jiri Mekyska (2)
(1) EUP Mataró (TCM – UPC), Avda. Ernest Lluch 32, 08302 MATARO (BARCELONA), SPAIN
(2) Brno University of Technology, Faculty of Electrical Engineering and Communication, Department of Telecommunications

font@tecnocampus.cat, faundez@eupmt.es, j.mekyska@phd.feec.vutbr.cz

*Abstract* –In this paper we present a method to identify people by means of thermal (TH) and visible (VIS) hand images acquired simultaneously with a TESTO 882-3 camera. In addition, we also present a new database specially acquired for this work.
The real challenge when dealing with TH images is the cold finger areas, which can be confused with the acquisition surface. This problem is solved by taking advantage of the VIS information.
We have performed different tests to show how TH and VIS images work in identification problems. Experimental results reveal that TH hand image is as suitable for biometric recognition systems as VIS hand images, and better results are obtained when combining this information. A Biometric Dispersion Matcher has been used as a feature vector dimensionality reduction technique as well as a classification task. Its selection criteria helps to reduce the length of the vectors used to perform identification up to a hundred measurements. Identification rates reach a maximum value of 98.3% under these conditions, when using a database of 104 people.

1. **INTRODUCTION**

The problem of people identification has been widely addressed [1, 2]. Different applications deal with reliable identification and/or verification approaches [3]. In this context biometric recognition turns to automatic identification of an individual by using certain personal attributes. These can be split into two groups: physiological, such as face, fingerprint, iris, hand and finger geometry and behavioral such us signature, gait or key stroking.

Generally, the performance of a biometric system is largely affected by some factors: quality of the sensor, conditions of the acquisition (such as lighting conditions in face recognition problems), feature extractor (length of the vector and discriminative power of each vector component), and classifier. No matter how well the process is conducted, the process is always subject to errors. If the variances between intra-class and inter-class are not different enough, there will be no separation at all and people will be misclassified. We have to be sure that discrepancy within an individual (different samples of the same person) is lesser than those from different individuals [4]. Thus, the scientific community has devoted a lot of effort to improve this issue. For instance, fingerprint, face, hand geometry among others have to be normalized using rotation and translation through standardized procedures. This is, in most situations, the most critical operation for the biometric system [5].

In this paper we present a novel approach for biometric recognition based on hand images obtained by a thermographic camera that contains a VIS and TH sensor. One major drawback to use this technology is its cost that could range from $10,000 to $50,000. However, this kind of technology is improving and we think that in the near future the price of these devices will be significantly lower. This research has been performed by a professional TESTO 882-3 camera. The main characteristics of this camera are its thermal image resolution: 320 x 240 pixels, its visible image resolution: 640 x 480 pixels and its thermal sensitivity (NETD) < 0.06 ºC at 30ºC. Fig. 1 shows the physical aspect of this camera.

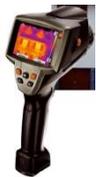
Fig. 1: Testo 882-3.

In this paper we present a straightforward method to overcome the major difficult aspect to deal with hand TH images. It is the segmentation of the image in order to keep the desired region of interest for extracting a feature vector. This process could be extraordinarily difficult because some individuals have cool fingers. These cool finger areas are very close to those from the acquisition table, being even difficult for a human being. There are some sophisticated methods to approach this kind of problem. Using simple methods like Otsu [6] gives poor quality. One of these more sophisticated methods, Active Shape Modelling (ASM) [7] does not provide good enough results when applied to cool finger hands. Texture segmentation based on semi-local region descriptor works pretty well in general images, but fails with cool areas [8]. In the cool areas there are not significant edges, but we have valuable and reliable information from the VIS image.

This paper presents a biometric identification system based on VIS and TH hand images alone as well as combined. The use of fusion methods improves results, reliability and robustness of the biometric system [9] up to 98.3% identification accuracy.

## 2. HAND IMAGE ACQUISITION

We have acquired a new hand image database with a thermal camera testo 882-3 that provides a visible and a thermal image. This database consists of 104 users acquired in five different sessions with a time slot of one week between them except the last one, which was acquired one month later. Four VIS plus four TH images have been acquired per person and acquisition session. Two of the images corresponds to the palm and the other two to the dorsum. In this paper, we have only used the dorsum. Fig. 2 shows an example of the acquired data for a specific user. It consists of the first five samples from VIS and TH that will be used to train our model. Five more samples, distributed in 5 different tests will be used to conduct the test.

It is important to point out that between consecutive images acquisition the hand is lifted, some hand exercise is done, and the hand is put down again on the acquisition surface. This procedure is followed in order to decorrelate consecutive acquisitions during the same day.

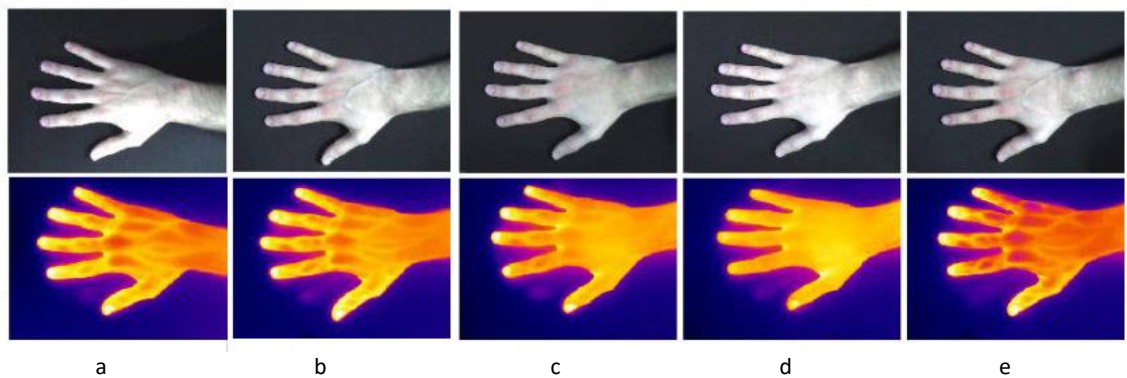

        a                  b                  c                  d                  e

Fig. 2: Example of images from the same person.

Humans can easily perform person identification with VIS faces, basically because they are used to doing that. With hands, it is different, because we are not used to recognizing each other this way, but we can guess that the samples shown in the first row of Fig. 2 have some common features. We can realize the big difference between the first row and second row from the same figure. TH hand images present a higher variability not seen in VIS images. Compare for example the samples d) and e) (see Fig. 2). Nevertheless the computer has different skills than a human being and this implies that we cannot forecast the automatic performance based on human performance.

The major challenge comes from the fact that many users present some kind of difficult issue (see i.e Fig. 3
and Fig. 4). It is hard to deal with sleeves (a) (casts shadows), rings (b), colored-nails (c) and (e) and cold fingers (c,d and e).

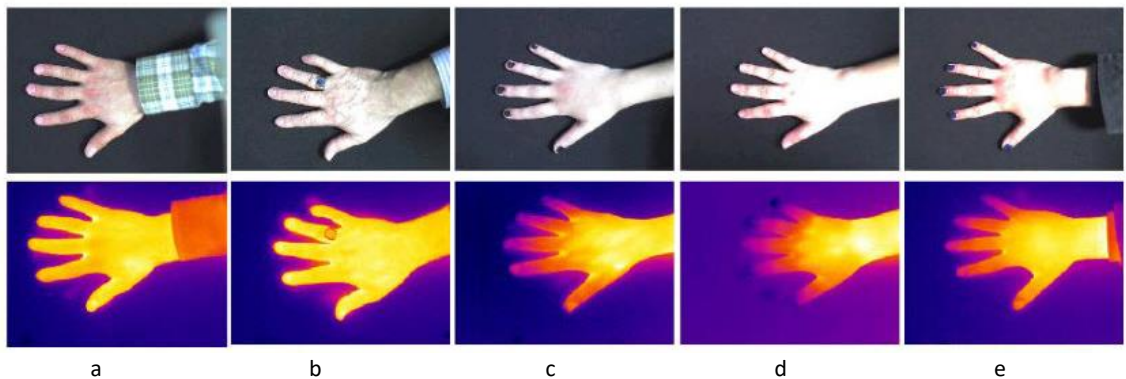

        a                  b                  c                  d                  e

Fig. 3: Hand images with relevant probems – I.

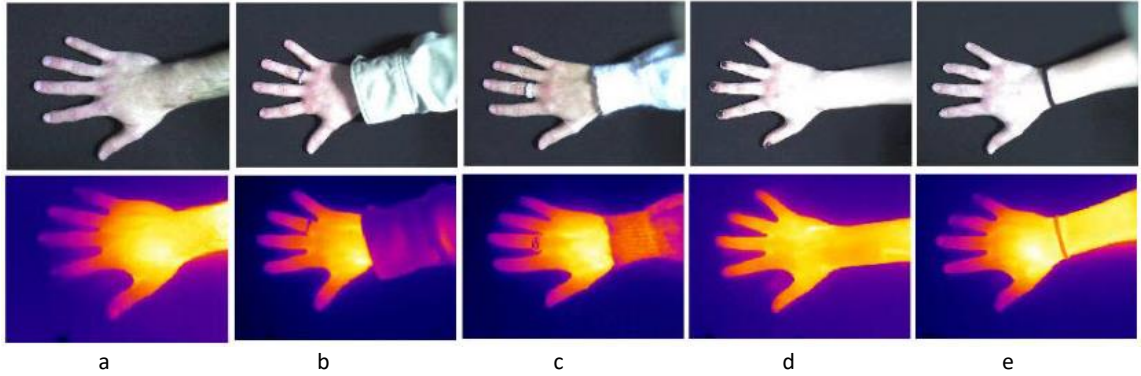

| a | b | c | d | e |

Fig. 4: Hand images with relevant problems – II.

Other problems can be found (see Fig. 4) as wrists (e) and for example rings placed on different fingers from the expected ring finger (c). Cold areas are present in this example in (a), (b), (c) and (e). Observe how colored-nails (d) work as a minor insulation surface.

### 3. SEGMENTATION METHOD

Image segmentation is typically used to locate objects and boundaries in images. More precisely, image segmentation for TH hand image is the process of assigning a label to every pixel in that image such that pixels with the same label share a desired characteristic (pixel from the user hand). While the segmentation process seems quite strightforward with warm hands (we see a clear heat pattern in 2nd row of Fig. 2) it is not clear when dealing hand images with cold fingers (see Fig. 3 and Fig. 4). Segmentation methods over this cold hands does not give good results because methods discard the cold areas. In order to avoid this drawback we propose a method that use the VIS segmented image. This process based on empirically found settings takes advantage of the thermographic camera, which provides two images at the same time: VIS and TH. With this information, it is possible to find the relation between these two images, i.e. find the mutual translation, rotation and scale.

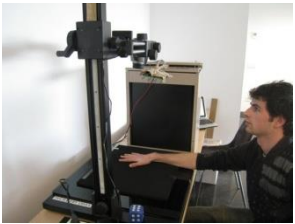

Fig. 5: Acquisition scenario.

This method works well if the camera has been placed in the same place during all the acquisition process, which is our case. Focal distance remains the same for all obtained users. Fig. 5 shows the acquisition setup.

The detailed steps to perform the TH segmentation are as follows:

1.- Firstly the VIS image was binarized using empirically found threshold (EF), by means of Otsu's method [6]. (see Fig. 6 and Fig. 7)

2.- The VIS binary mask was transformed to the TH mask using the correct rotation, translation and scaling parameters. (see Fig. 8 and Fig. 9)

Step two can be made empirically, because lab conditions and the camera was fixed during all acquisition sessions or by means of a more universal method based on the image registration. In this case to find the best rotation, translation and scaling settings we use an optimization function based on the simplex search method [10]. In our conditions the first method outperformed the second one.

As soon as TH images present cool areas the goodness of this method comes to light (see Fig. 10 and Fig. 11).

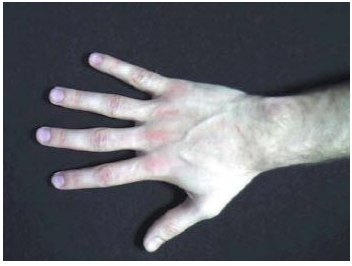
Fig. 6: Original VIS image.

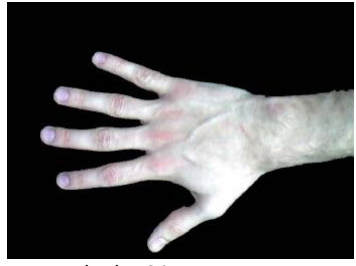
Fig. 7: Masked VIS image.

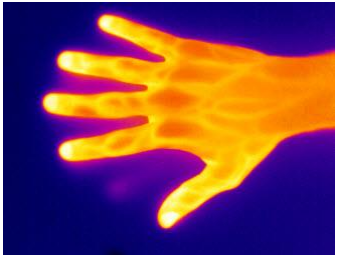
Fig. 8: Original TH image.

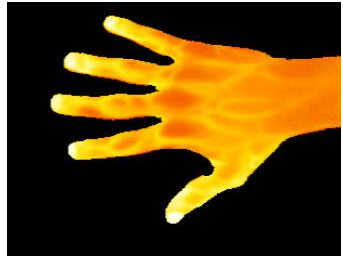
Fig. 9: Masked TH image.

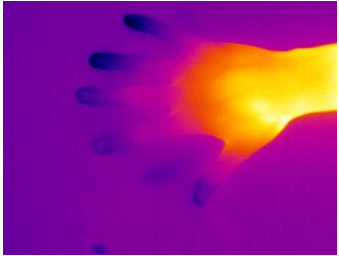
Fig. 10: TH image with cool areas.

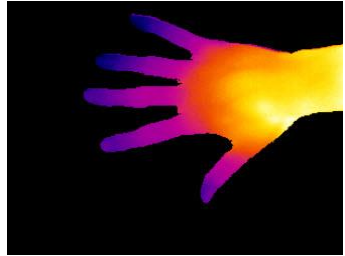
Fig. 11: TH image correctly Segmented.

It is important to point out that the main relevant information present in TH images that is not contained in VIS ones is the heat pattern distribution related with the vascular pattern (commonly referred to as vein pattern) on the hand surface rather than the hand contour. Thus, this is the proposed procedure to extract this information removing the background information.

### 4. FEATURE EXTRACTION AND RECOGNITION

Discrete Cosine Transform (DCT) has emerged as the de-facto image transformation in most visual systems. DCT has been widely deployed by modern video coding standards, for example, MPEG, JVT etc. Its main properties are decorrelation, energy compactation, separability, symetry and orthogonality.

$$C(u,v) = \alpha(u)\alpha(v) \sum_{x=0}^{N-1} \sum_{y=0}^{N-1} f(x,y) \cos\left[\frac{\pi(2x+1)u}{2N}\right] \cos\left[\frac{\pi(2y+1)v}{2N}\right]$$

For $u,v = 0, 1, 2, ..., N-1$ and $\alpha(u)$ and $\alpha(v)$ defined as

$$\alpha(i) = \begin{cases} \sqrt{\dfrac{1}{N}} & \text{for } i = 0 \\ \sqrt{\dfrac{2}{N}} & \text{for } i \neq 0 \end{cases}$$

In order to deploy feature extraction the first step has been to perform a two dimensional Discrete Cosine Transformation (DCT) [11] which provides an image of the same size but it has a strong energy compaction (energy compacted in the low frequency bands). The original signal is converted to the frequency domain thanks to cosine functions. Usually the lowest frequencies (first vector components) carries the majority of the relevance. The higher frequencies will be discarded because represent fine image details and noise as well, and are not suitable for person identification. In our experiments we have selected up to 500 coefficients (see experimental results section).

The second step is related to recognition process. Biometric Dispersion Matcher (BDM) is a simple classification method linked to Discriminant Analysis. Works under the construction of two groups:

- Equal group: where we have pairs of measurements of a characteristic related to the same person.
- Unequal group: where we have pairs of measurements of a characteristic related to different persons.

For an easy exposition, suppose that we deal with a feature vector of one component. For each group we will have a distribution of differences between pairs. These distributions under the usual Gaussian hypothesis give us the following expressions:

$$p(x \mid U) = N\left(x \mid 0, 2\left(\sigma_p^2 + \sigma_i^2\right)\right)$$
$$p(x \mid E) = N\left(x \mid 0, 2\sigma_i^2\right)$$

where $\sigma_i$ is the variance of the vector component over the same user (over its samples), and $\sigma_p$ is the variance of that component, over the whole user population. For the unequal group we have a zero mean. Variance is equal to two times the variance sum of both variances. For the equal group, zero mean and variance is equal to two times the variance of the component of the same individual.

The expression $\sigma_i^2 / \left(\sigma_p^2 + \sigma_i^2\right)$ is related to the discriminative power of the characteristic component under consideration. BDM use this expression to select the best feature vector components. It is important to point out the way BDM will conduct the selection criteria. Through a σ threshold all components greater than this threshold will be discarded.

The method performs the classification applying a discriminative function **Error! Reference source not found.** that lets us define an easy classification rule.

$$g(x) = \ln\left\{\frac{p(x \mid U)}{p(x \mid E)}\right\} + \ln\left\{\frac{p(U)}{p(E)}\right\} \quad (1)$$

Then the decision is straightforward:

$$\begin{cases} U & \text{if } g(x) > 0 \\ E & \text{otherwise} \end{cases}$$

When dealing with the general case the steps required are as follows:

1. Compute the DCT2 and select its 100 first components
2. Compute the covariance matrix from grups *E* and *U*
3. Compute the ratio between both covariance (*E / U*)
4. Sort this ratio and select the best components (those with a value lower than σ-threshold)
5. Compute the discriminative function *g(x)*

## 5. EXPERIMENTAL RESULTS

In order to quantify both types of hand information coming from VIS and TH images we have processed the following hand regions: index finger (alone), central hand zone (without fingers) and the whole normalized hand [13]. Visual direct information (**Error! Reference source not found.**) has been compared with hand texture information obtained from Principal Component Analysis (PCA). These two methods are called "direct" and "texture" in the next results tables. The idea behind PCA is to enhance texture variance due to palm prints (see **Error! Reference source not found.**).

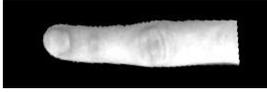
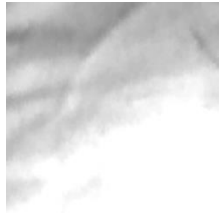
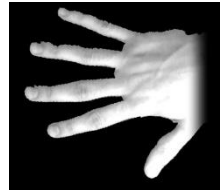

Fig. 12: VIS direct images, Finger, Background, Hand.

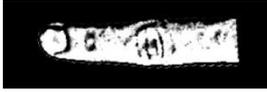
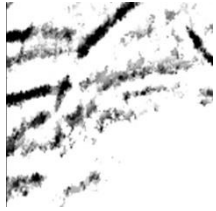
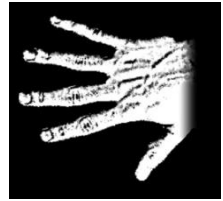

Fig. 13: VIS TEXTURE images, Finger, Background, Hand.

### 5.1 VIS RESULTS

Visual hand information provides recognition performance close to the best performance in the literature. The results coming from the index finger are certainly surprising and promising, because segmentation and normalization are more easily normalized than the whole hand. We can see the identification recognition rates in Tab. 1 when the dimension of the feature vector is equal to 100 components. Identification rates are presented for each testing image (test 1 corresponds to the second image of session 3, test 2 is the first image of session 4, and so on). This vector comes from the feature extractor, which is the DCT2 in our study, as we have discussed in the previous section.

|  | ID % | Test 1 | Test 2 | Test 3 | Test 4 | Test 5 | Mean | Std Dev |
|---|---|---|---|---|---|---|---|---|
| Finger | Direct | 99.0 | 99.0 | 94.2 | 98.1 | 100.0 | 98.1 | 2.3 |
|  | Texture | 99.0 | 99.0 | 93.3 | 94.2 | 93.3 | 95.8 | 3.0 |
| Background | Direct | 97.1 | 93.3 | 88.5 | 87.5 | 95.2 | 92.3 | 4.2 |
|  | Texture | 97.1 | 94.2 | 89.4 | 85.6 | 92.3 | 91.7 | 4.4 |
| Hand | Direct | 99.0 | 99.0 | 96.2 | 98.1 | 99.0 | 98.3 | 1.3 |
|  | Texture | 99.0 | 99.0 | 96.2 | 96.2 | 98.1 | 97.7 | 1.5 |

Tab. 1: Identification Rate (%) for VIS images.

In general the texture approach gives worse results than the direct visual image (containing the background, which means that no hand segmentation is performed). And the normalized full hand gives the best performance over the different options tested.

It is possible to appreciate how BDM selects the desired characteristics once the feature vector length is increased from 100 to 200, 300, 400 and 500 components.

|  | RealDim | 100 | 200 | 300 | 400 | 500 |
|---|---|---|---|---|---|---|
| Finger | Direct | 77 | 73 | 83 | 95 | 92 |
|  | Texture | 95 | 114 | 153 | 90 | 150 |
| Background | Direct | 80 | 98 | 106 | 120 | 71 |
|  | Texture | 80 | 99 | 67 | 76 | 71 |
| Hand | Direct | 99 | 81 | 60 | 135 | 102 |
|  | Texture | 96 | 164 | 189 | 113 | 141 |

Tab. 2: VIS Selected components from BDM.

|  | Threshold | 100 | 200 | 300 | 400 | 500 |
|---|---|---|---|---|---|---|
| Finger | Direct | 0.60 | 0.40 | 0.37 | 0.37 | 0.36 |
|  | Texture | 0.73 | 0.52 | 0.52 | 0.46 | 0.50 |
| Background | Direct | 0.81 | 0.75 | 0.75 | 0.76 | 0.70 |
|  | Texture | 0.82 | 0.76 | 0.72 | 0.72 | 0.72 |
| Hand | Direct | 0.65 | 0.37 | 0.34 | 0.37 | 0.34 |
|  | Texture | 0.55 | 0.48 | 0.44 | 0.37 | 0.38 |

Tab. 3: VIS Threshold for $\sigma$.

The σ threshold decreased in order to reduce the number of selected components from the DCT2. In the case of finger analysis, we go from a threshold of 0.60 to select 77/100 components to a threshold of 0.36 to select 92/500 components.

The feature selection uses a natural approach when using BDM, and basically discards those components of the DCT vector that carry little useful information. In Tab. 2 and Tab. 3 we can appreciate how the method decreases the threshold in order to discard additional components. For DCT vectors with high dimension the BDM builds a singular covariance matrix. One way to mitigate these effects is with the feature selection.

The performance of the BDM across all the characteristics vector lengths is displayed in Fig. 14.

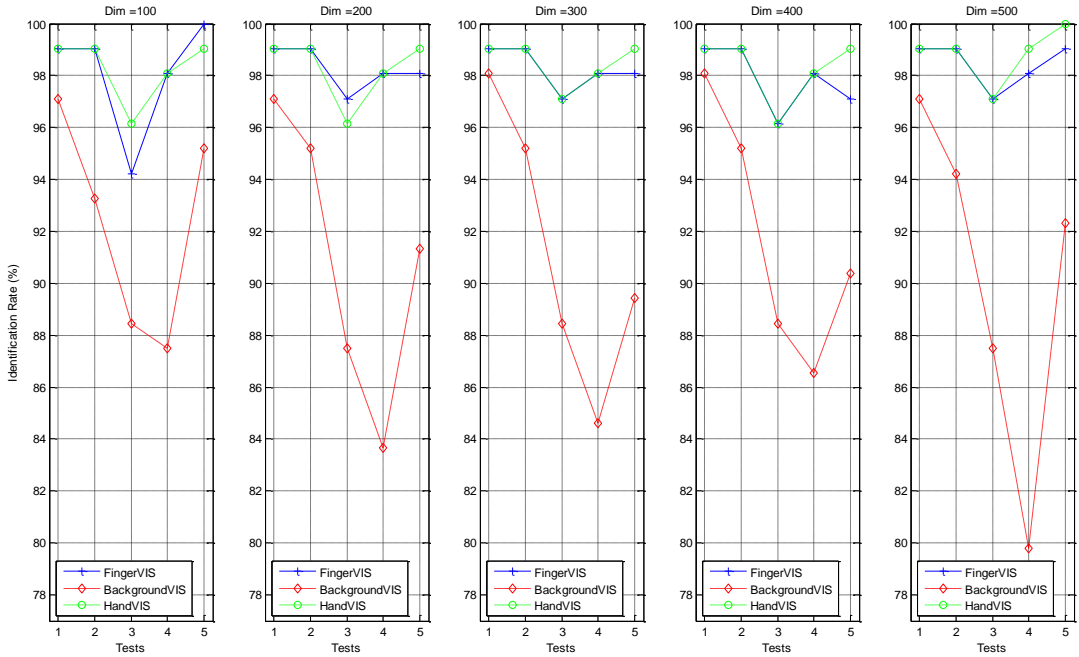

Fig. 14: VIS performance through increased characteristic vector (100, 200, 300, 400, 500).

### 5.2 TH RESULTS

The results coming from TH images are shown in Tab. 4. It is worth observing that the best results come from Test 1. This is obvious because this test was acquired during the same day of the last data training sample. The environmental conditions (i.e. temperature) were the same as well as the conditions of the user. However, as we move forward within tests, all biometrics characteristics decrease its performance. This phenomena is linked with the acquisitions of TH images. While TH images are not affected by lighting conditions they are affected by the user condition and environment (temperature). The heat hand pattern during winter is different from the one coming from summer. Further, if the user comes to the acquisition session once he has trained in a gym or if he just came right after waking up in the morning, the heat pattern can be different.

It is worth mentioning that the texture method on finger TH images does not provide good enough results. For this reason, it has not been included in the next tables.

|  | ID % | Test 1 | Test 2 | Test 3 | Test 4 | Test 5 | Mean | Std Dev |
|---|---|---|---|---|---|---|---|---|
| Finger | Direct | 95.2 | 49.0 | 40.4 | 49.0 | 44.2 | 55.6 | 22.4 |
|  | Texture |  |  |  |  |  |  |  |
| Background | Direct | 92.3 | 63.5 | 45.2 | 44.2 | 34.6 | 56.0 | 22.8 |
|  | Texture | 75.0 | 40.4 | 24.0 | 24.0 | 17.3 | 36.2 | 23.3 |
| Hand | Direct | 98.1 | 91.3 | 81.7 | 76.9 | 76.9 | 85.0 | 9.4 |
|  | Texture | 96.2 | 78.8 | 69.2 | 71.2 | 73.1 | 77.7 | 10.9 |

Tab. 4 : Identification Rate (%) for TH.

While the finger in VIS spectrum performs good results across all tests, in TH it decreases drastically once we move forward through test 2, test 3 and so on. This is because there is too much variation that cannot be explained alone with the index TH finger. Comparing these results to those coming from the whole hand we observe that there is a performance drop but not as strong as with the finger.

Experiments show how the finger and hand in TH behave. Increasing the DCT2 vector of the TH finger from 100 to 500 does not imply any improvement, and the selected components stay at 58 components (see Tab. 5). With the TH hand there is an increase in the number of selected components, from 63 to 115 when moving from 100 to 500. This means that there is additional relevant information to take into account for recognition purposes beyond the first 100 components.

|  | Real Dim | 100 | 200 | 300 | 400 | 500 |
|---|---|---|---|---|---|---|
| Finger | Direct | 49 | 85 | 53 | 54 | 58 |
| Background | Direct | 80 | 84 | 62 | 101 | 74 |
|  | Texture | 86 | 86 | 78 | 50 | 88 |
| Hand | Direct | 63 | 87 | 98 | 136 | 115 |
|  | Texture | 50 | 50 | 84 | 78 | 86 |

Tab. 5: TH Selected Components for BDM.

|  | Threshold | 100 | 200 | 300 | 400 | 500 |
|---|---|---|---|---|---|---|
| Finger | Direct | 0.75 | 0.71 | 0.58 | 0.53 | 0.50 |
| Background | Direct | 0.95 | 0.92 | 0.89 | 0.92 | 0.90 |
|  | Texture | 0.95 | 0.91 | 0.91 | 0.88 | 0.91 |
| Hand | Direct | 0.68 | 0.57 | 0.51 | 0.53 | 0.47 |
|  | Texture | 0.55 | 0.42 | 0.45 | 0.41 | 0.41 |

Tab. 6: TH Threshold for $\sigma$.

The performance of the whole hand overpasses the other measurements. It has been proved that the use of a finger does not provide too much relevant information in the TH spectrum (Fig. 15). The strategy to overcome this usefulness of finger results is comparing finger temperatures with background temperatures. These differences can fix some variability and improve results.

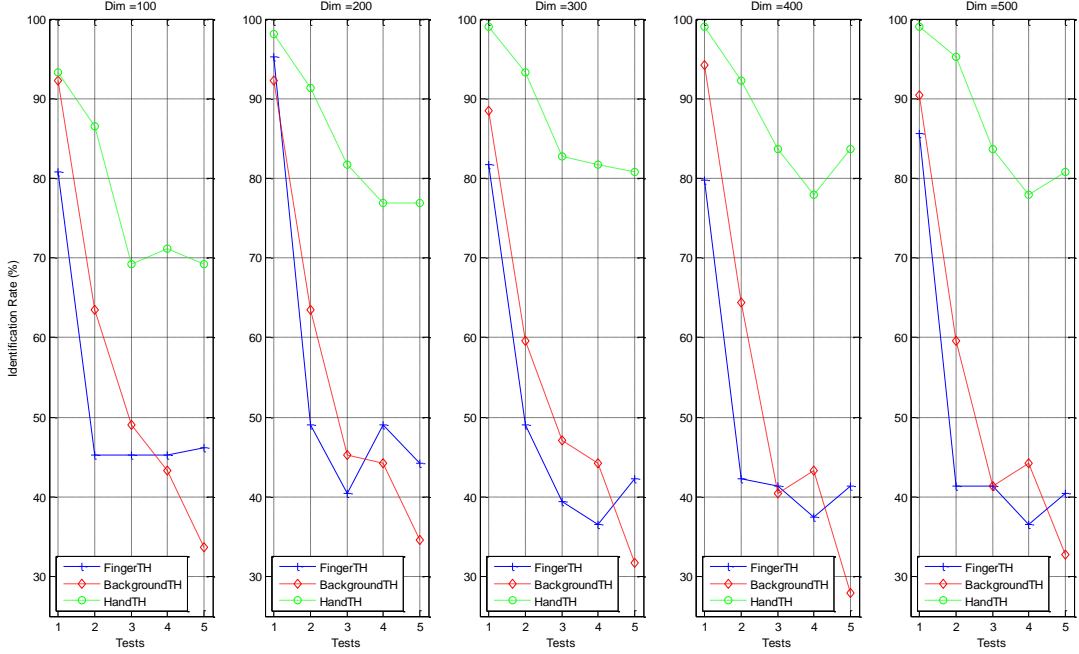

Fig. 15: TH performance through increased characteristic vector (100, 200, 300, 400, 500).

### 5.3 VIS AND TH RESULTS

One reasonable way to improve performance is through fusion [14, 15]. Consider the general problem where we have an input pattern $X_1$ (a VIS image), that becomes a feature vector $x_1$ (selected DCT2 coefficients from $X_1$) and applied to a biometric system $S_1$ (BDM) to obtain one of the $c$ possible classes $\{w_1, ..., w_c\}$. Consider another input pattern $X_2$ (a TH image), that becomes a feature vector $x_2$ and applied to a biometric system $S_2$ (BDM again) to obtain $c$ possible classes. We could go on with more input patterns, and saying we have S different systems.

From a Bayesian decision theory the input pattern $X$ should be assigned to the class $w$ that has the highest posterior probability.

$$if\ P(w_k \mid x_1, \cdots, x_S) \geq P(w_i \mid x_1, \cdots, x_S)\ \forall i \in S \Rightarrow X \to w_k \qquad (2)$$

The posterior probabilities in (2) can be expressed in terms of the conditional joint probability using Bayes rule. If we assume independence across the $S$ feature vectors, then it is possible to express the joint probability density as a product of marginal conditional densities (3).

$$p(x_1, \cdots, x_S \mid w_i) = \prod_{j=1}^{S} p(x_j \mid w_i) \qquad (3)$$

Three different strategies have been used.

- The first one comes from decision level fusion, majority voting. In our case this strategy does not provide good enough results, because we combine information from two different information sources, say VIS and TH. This strategy is suitable when the number of biometric decision system is much larger. Let's say we have k different biometric decision system, then the decision taken is from the class with most votes.

- The second one comes from score decision fusion. In this case: product, mean, median, max and min rules have been applied. Assuming in all cases equal prior probabilities, which is our case, the rules are as follows:

Results across the different vector lengths are shown in Fig. 16. Again the best fusion performance comes from hand, and quite close to the index finger results. With this outstanding performance an increase in the feature vector from 100 to 500 does not change the results.

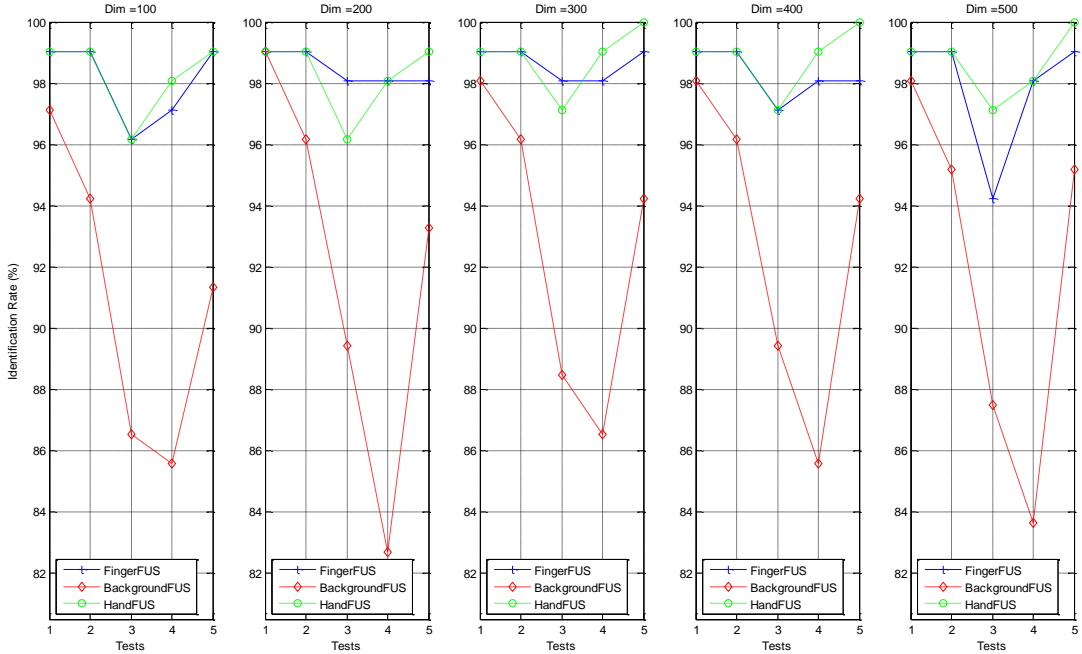

Fig. 16: VIS + TH performance through increased characteristic vector (100, 200, 300, 400, 500).

We summarize the performance results (see **Error! Reference source not found.**) coherent with the visual information shown in Fig. 16

| | ID % | Test 1 | Test 2 | Test 3 | Test 4 | Test 5 | Mean | Std Dev |
|---|---|---|---|---|---|---|---|---|
| Finger | Normal | 99.0 | 99.0 | 96.2 | 97.1 | 99.0 | 98.1 | 1.4 |
| Background | Normal | 97.1 | 94.2 | 86.5 | 85.6 | 91.3 | 91.0 | 4.9 |
| Hand | Normal | 99.0 | 99.0 | 96.2 | 98.1 | 99.0 | 98.3 | 1.3 |

Tab. 7: Identification Rate (%) from VIS + TH.

- The third fusion strategy comes from trained rules: Some classifiers should have more relevance on the final result. This is achieved by means of some weighting factors that are computed in the following way:

$$Comb = \alpha \, VIS + (1-\alpha) \, TH. \qquad (4)$$

The weighting factor α in (4) can be interpreted in an interesting way when combining two classifiers. If *α=0* then the *Comb* results stand for *TH* and the *VIS* effect is fully discarded. With *α=1* the results come from *VIS* and nothing from *TH*.

We have shown results from the background (Fig 17**Error! Reference source not found.**). The results from the hand have been discarded because they were similar to those obtained by the finger. Some interesting remarks about these results:
Analysis of the finger:
- o The performance from *α=0* to *α=1* begins with a low performance (realted with TH) that increases drastically as *α* increase.
- o The performance curve seems to follow a monotonic growth curve that goes from the lowest TH performance to the highest VIS performance. The exception is in test 5 where the maximum is obtained with *α=0.35*.
- o The different curves represented in the performace graphs show the effect of combining the direct score from BDM (best performance) with two normalization schemes, one applying the usual reduction and translation $N = \dfrac{y - \bar{y}}{s_y}$ (worst results), and the other following a minimax normalization score (which guarantees values between 0 and 1)

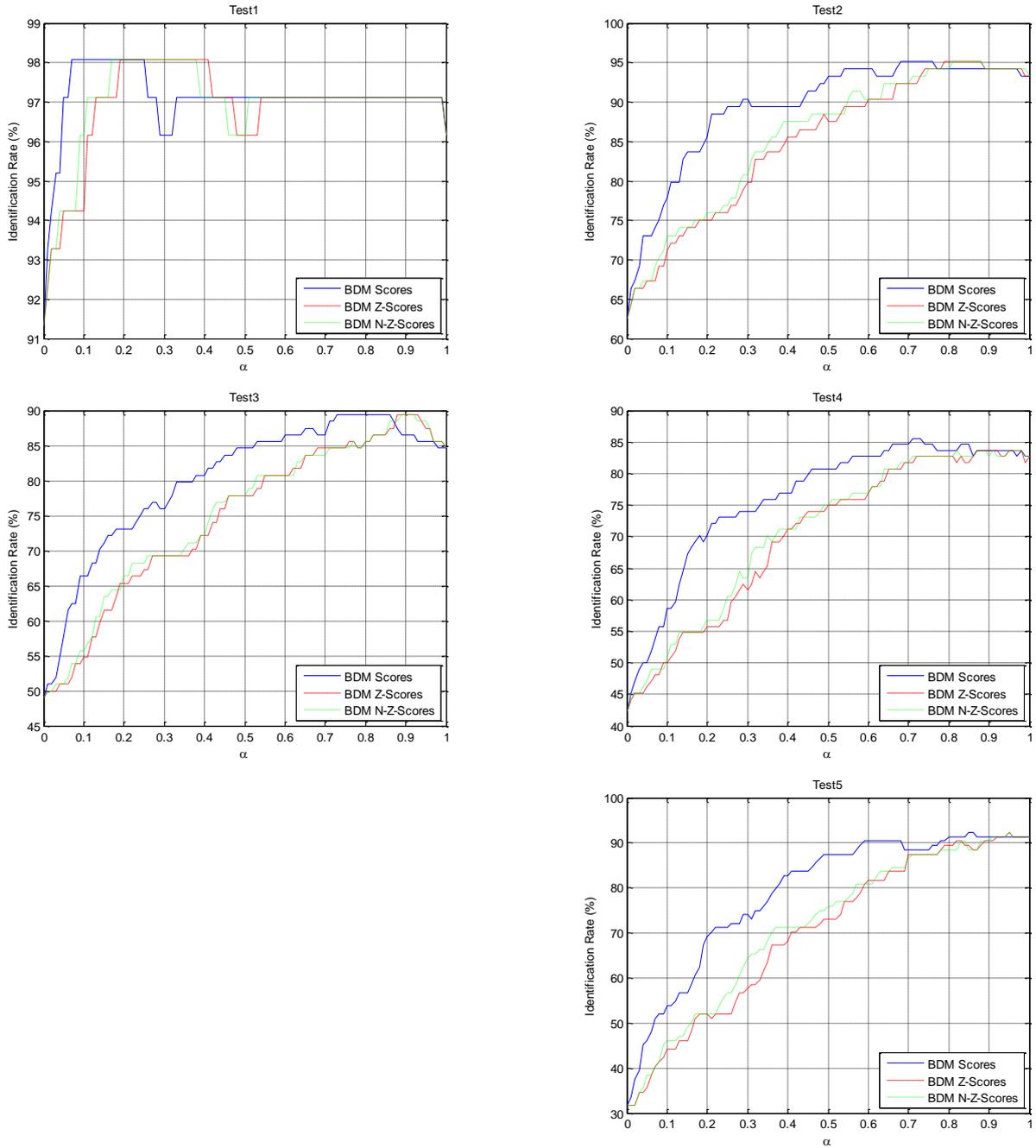

Fig 17: Trained rule combining Background TH and VIS

Analysis of the background:
- o The performance from *α=0* to *α=1* begins with a low performance (realted with *TH*) that increases smoothly as *α* increases. (exception in test 1 because it is a test close to the data train set)
- o The best performance is not reached at the *VIS* side (as with the finger). It is reached in some optimum *α* value (0.7 to 0.8).
- o Again the best performance comes from combining direct scores.

## 6. CONCLUSIONS

We have seen how to avoid the cold finger areas in order to get a better segmented TH image. In fact these approaches are only necessary when temperatures from the finger are close to the surface. Once the TH image is well segmented we have observed different performance.

We have presented a new approach combining VIS and TH spectrum for hand-based identification recognition. Information from VIS provides identification rates close to 99% while from TH close to 85% in the case of the hand and 56% in case of the finger.

If we can afford a recognition through VIS finger it is not advisable to use TH finger image. However, using a better camera, which on the one hand improves sensitivity, and on the other hand resolution, may be able to get some finger vein pattern that we did not observe with our camera.

The TH image of the whole hand improves substantially the results from the finger. These results are, still not close to visual but are good enough to be considered a complement to VIS images. The reason is obvious: a VIS hand image could be easily changed with a fake (imagine a simple photograph), while with the TH image it is quite more difficult. The use of both spectrum hand images (TH and VIS) could help to prevent false intrusion and improve security reliability. That is clear, because providing a fake physical biometric design to circumvent the biometric system can be relatively easy with VIS systems. The use of an another image from the TH spectrum leaves the system with a more robust design against counterfeiting.

## 7. ACKNOWLEDGEMENTS

This work has been supported by FEDER, MEC, TEC2009-14123-C04-04, KONTAKT-ME 10123 and VG20102014033.

## 8. REFERENCES

[1] C. M. Bishop, *Pattern recognition and machine learning*, New York: Springer, 2006.
[2] R. O. Duda, P. E. Hart, and D. G. Stork, *Pattern classification*, 2nd ed., New York: Wiley, 2001.
[3] L. O'Gorman, "Comparing passwords, tokens and biometrics for user authentication," *Proceedings of the IEEE,* vol. 91, no. 12, pp. 2021-2040, 2003.
[4] J. Fàbregas, and M. Faundez-Zanuy, "Biometric dispersion matcher," *Pattern Recognition,* vol. 41, no. 11, pp. 3412-3426, 2008.
[5] E. Yörük, E. Konukoglu, B. Sankur *et al.*, "Shape-Based hand Recognition," *IEEE Trans. on Image Processing,* vol. 15, no. 7, pp. 1803-1815, 2006.
[6] N. Otsu, "A Threshold Selection Method from Gray-Level Histograms," *IEEE Transactions on Systems, Man, and Cybernetics,* vol. 9, no. 1, pp. 62-67, 1979.
[7] e. a. Ginneken B., "Active Shape Model Segmentation with Optimal Features," *IEEE Transactions on Medical Imaging* 2002.
[8] J.-P. T. a. X. B. Nawal Houhou, "Fast Texture Segmentation Based on Semi-Local Region Descriptor and Active Contour," *Numer. Math. Theor. Meth. Appl.,* vol. 2, no. 4, pp. 445-468, 2009.
[9] A. Jain, K. Nandakumar, and A. Ross, "Score normalization in multimodal biometric systems," *Pattern Recognition,* vol. 38, no. 12, pp. 2270-2285, 2005.
[10] J. C. Lagarias, J. A. Reeds, M. H. Wright, and P. E. Wright, "Convergence Properties of the Nelder-Mead Simplex Method in Low Dimensions," *SIAM Journal of Optimization,* vol. 9, no. 1, pp. 112-147, 1998.
[11] A. B. Watson, "Image Compression Using the Discrete Cosine Transform," *Mathematica Journal,* vol. 4, no. 1, pp. 81-88, 1994.
[12] J. Fàbregas, and M. Faundez-Zanuy, "Biometric dispersion matcher versus LDA," *Pattern Recognition,* vol. 42, no. 9, pp. 1816-1823, 2009.
[13] H. D. Erdem Yörük, Bülent Sankur, "Hand Biometrics," *Image and Vision Computing,* vol. 24, pp. 483-497, 2006.
[14] A. Ross, K. Nandakumar, and A. K. Jain, *Handbook of multibiometrics*: Springer, 2006.
[15] L. I. Kuncheva, *Combining Pattern Classifiers*: Wiley-Interscience, 2004.